\documentclass[a4paper, 10pt, conference]{ieeeconf}       
\usepackage{FG2020}

\FGfinalcopy 

\IEEEoverridecommandlockouts                              
\usepackage{graphics} 
\usepackage{epsfig} 
\usepackage{mathptmx} 
\usepackage{times} 
\usepackage{amsmath} 
\usepackage{amssymb}  
\let\labelindent\relax
\usepackage{enumitem}
\usepackage{multirow}
\usepackage{todonotes}
\usepackage{hyperref}
\hypersetup{
    colorlinks=true,
    linkcolor=red,
    filecolor=magenta,      
    urlcolor=red,
}

\def\FGPaperID{8} 

\title{\LARGE \bf
Real-time Facial Expression Recognition ``In The Wild'' by Disentangling 3D Expression from Identity
}

\author{\parbox{16cm}{\centering
    {\large Mohammad Rami Koujan$^{1}$, Luma Alharbawee$^{1}$, Giorgos Giannakakis$^{2}$, Nicolas Pugeault$^{1}$, Anastasios Roussos$^{2}$}\\
    {\normalsize
    $^1$ College of Engineering, Mathematics and Physical Sciences, University of Exeter, UK\\
    $^2$Institute of Computer Science (ICS), Foundation for Research and Technology - Hellas (FORTH), Greece}}
}

\begin{document}

\ifFGfinal
\thispagestyle{empty}
\pagestyle{empty}
\else
\author{Anonymous FG2020 submission\\ Paper ID \FGPaperID \\}
\pagestyle{plain}
\fi
\maketitle

\begin{abstract}
Human emotions analysis has been the focus of many studies, especially in the field of 
Affective Computing, and is important for many applications, e.g. human-computer
intelligent interaction, stress analysis, interactive games, animations, etc. 
Solutions for automatic emotion analysis have also benefited from the development of deep learning approaches and 
the availability of vast amount of visual facial data on the internet. 
This paper proposes a novel method for human emotion recognition from a single RGB image. We construct a large-scale dataset of facial videos (\textbf{FaceVid}), rich in facial dynamics, identities, expressions, appearance and 3D pose variations. We use this dataset to train a deep Convolutional Neural Network for estimating expression parameters of a 3D Morphable Model and combine it with an effective back-end emotion classifier. 
Our proposed framework runs at 50 frames per second and is capable of 
robustly estimating parameters of 3D expression variation   
and accurately recognizing facial expressions from in-the-wild images. 
%
%
We present extensive experimental evaluation that shows that the proposed method outperforms the compared techniques in estimating the 3D expression parameters and achieves state-of-the-art performance in 
recognising the basic emotions from facial images, as well as recognising stress from facial videos. 

\end{abstract}


\section{Introduction}
\label{sec:intro}
As the Computer Vision (CV) and Machine Learning (ML) fields advance, the study of human faces progressively receives a notable attention due to its central role in a plethora of key applications. Human emotion recognition is an increasingly popular line of research, around which many research studies revolve. The aim of most of these studies is to automate the process of recognising a human's emotion from a captured image. Solving this problem successfully is immensely beneficial for a myriad of applications, e.g. human-computer intelligent interaction, stress analysis, interactive computer games, emotions transfer, most of which have been the focus of the Affective Computing field.

The recent availability of large benchmarks of facial expression (images and videos) and the fast development of deep learning approaches has lead to high performance in facial expression recognition for image data captured in both controlled and unconstrained conditions (``in the wild''). 


This paper presents a novel approach for recognizing human emotions from a \textit{single} facial image. The proposed approach capitalises on the recent advancements in 3D face reconstruction from monocular videos and Convolutional Neural Networks (CNNs) architectures that proved effective in the CV field. Our method is driven by the idea of disentangling the subject's expression from identity with the aid of \textit{3D Morphable Models (3DMM)} \cite{blanz1999morphable}. Given a \textit{single} image, we regress a vector representing the \textit{3D expression} of the depicted subject with the help of a novel Deep Convolutional Neural Network (DCNN), which we call \textit{DeepExp3D}. This expression vector is ideal as a feature vector since it achieves various invariances (with respect to the individual's facial anatomy, 3D head pose and illumination conditions), and we show that an emotion classifier trained on this feature can recognise expressions reliably and robustly. 
Our contributions in this work can be summarized as\footnote[1]{Project page: \url{https://github.com/mrkoujan/FER}}:
\begin{itemize}[leftmargin=*]
\item Collection and annotation of a new large-scale dataset of human facial videos (6,000 in total), which we call \textbf{FaceVid}. With the help of an accurate model-based approach that we propose to use during training, each video is annotated with the per-frame: 1)  facial landmarks, 2) 3D facial shape composed additively of identity and expression parts, 3) relative 3D pose of the head with respect to the camera.  
\item A robust deep convolutional neural network (CNN), termed as \textbf{DeepExp3D}, for regressing the expression parameters of a 3D Morphable Model of the facial shape from a single input image. Our network is robust to occlusions, illumination and view angle changes, and regresses the expression independently of the person's identity.
\item We connect DeepExp3D with a classification module for the \textbf{Facial Expression Recognition} (FER) task from the estimated expression vectors, leading to an integrated framework for the robust recognition of facial expressions from single images. 

\end{itemize}
Our trained DeepExp3D: \textbf{1)} outperforms state-of-the-art 3D face reconstruction methods in estimating the facial expression parameters, \textbf{2)} achieves state-of-the-art performance in FER from images and stress recognition from videos, \textbf{3)} can also be incorporated in other frameworks seeking to, e.g., recover the 3D geometry of facial image, image-to-image translation, facial reenactment, etc. 



\begin{figure*}[t!]
    \centering
    \includegraphics[width= 17cm, height=7.6cm]{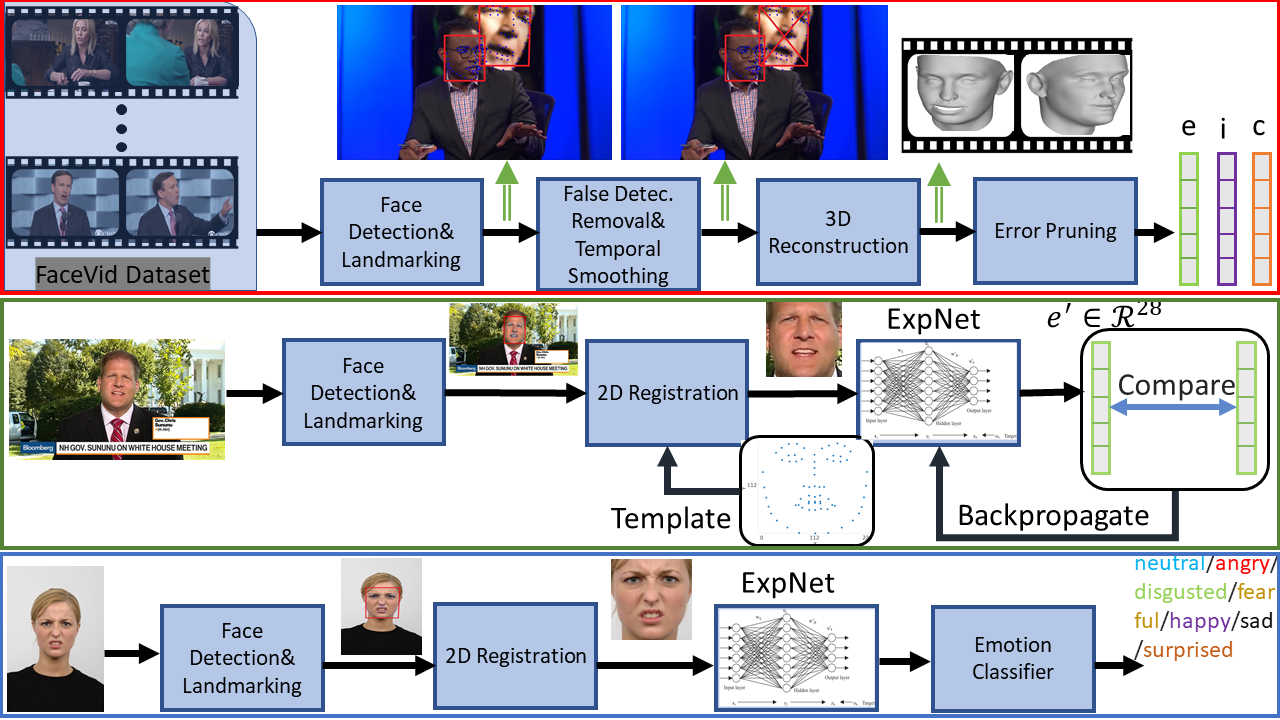}
    \caption{Proposed framework of Facial Expression Recognition (FER) from images. Top: \text{FaceVid} annotation process (sec. \ref{subsec: 3DMM} and \ref{subsec: dataset}). Middle: training of \textbf{DeepExp3D} (sec. \ref{subsec: ExpNet}). Bottom: final framework for FER (sec. \ref{subsec: EmotionClassifier}). Vectors $\mathbf{e}$, $\mathbf{i}$, $\mathbf{c}$ estimated in the annotation process (top) represent facial expression, identity and camera parameters, respectively.}
    \label{fig:framework}
\end{figure*}

\section{Related work}
\label{sec:Related work}

There is a large body of work tackling the seven-class problem of static facial expression recognition defined by Ekman and Friesen 
\cite{ekman1993facial}.
Current facial expression recognition methods divide approximately into two groups: traditional handcrafted methods (appearance, geometric, dynamic and fusion) and deep learning models. Handcrafted methods have been widely adopted for FER and rely on features \cite{jiang2016robust, dhall2018emotiw}. Nevertheless, they have shown their restrictions in practical applications \cite{koujan2018multi, leonovych2018texture}. Lately, deep learning, especially Convolutional Neural Networks (CNN), methods have proved competitive in many vision tasks, e.g, image classification, segmentation, emotion recognition, etc.

Xiao et al. \cite{xiao2016facial} tackle the poor generalization of deep neural networks when enough data is not available by combining region of interest (ROI) and K-nearest neighbors (KNN) for facial expression classification. In \cite{barros2017emotion}, an attention model composed of a deep CNN learns the location of emotional expression in a cluttered scene, leading to an improved facial expression recognition. Liu et al. \cite{liu2014deeply} proposed a deep learning approach trained on a geometric model of facial regions for facial expression analysis. Tang \cite{tang2013deep} proposed a CNN backed with a linear support vector machine (SVM) at the output and achieved the first place on the FER-2013 Challenge \cite{goodfellow2013challenges}. Liu et al. \cite{liu2014combining} suggested a facial expression recognition framework with 3D CNN and deformable action parts constraints to jointly localize specific facial action parts and recognize facial expressions. Peng \cite{peng2017reconstruction} focused on a synthesis CNN to produce a non-frontal view from a single frontal face and Richardson et al.  \cite{richardson20163d} transferred the face geometry from its image directly via a CNN based approach. The authors of \cite{otberdout2018deep} encode deep convolutional neural networks (DCNN) features with covariance matrices for facial expression recognition. In their paper, they show that covariance descriptors computed on DCNN features are more efficient than the standard classification with fully connected layers and softmax. For systematic and exhaustive surveys on automatic FER, we refer the reader to \cite{fasel2003automatic, li2018deep}. 

In contrast to these approaches, we propose to estimate an intermediate 3D-based representation of ``pure'' facial expression that is \textit{invariant} to all other parameters that contribute to the formation of the input image (shape and appearance variation related to the subject's identity, relative 3D pose variation, occlusions, strong illumination variations and other challenges of in-the-wild images). This means that, in contrast to the standard practice of most Deep Learning approaches in CV, we are not seeking to solve our problem (FER from single RGB images) in an ``end-to-end'' fashion; this would require a vast number of manually-annotated images in the level of facial expression classes, which would be laborious and prone to human annotation errors. Instead, we construct a large-scale video dataset and annotate all video frames individually with the expression parameters of a 3DMM. We are able to reliably automate this annotation process using an approach of 3D face reconstruction from videos that achieves high accuracy by exploiting the rich dynamic information that facial videos have. Using this dataset to learn to regress expression parameters from single RGB images, we massively simplify the problem of FER, since we use the expressions parameters as the features that feed our emotion classifier. These expression parameters are low-dimensional (28 dimensions) and exhibit a very wide range of invariance properties, therefore a classifier can be trained to recognise emotion classes with significantly less annotated data. This approach leads to a robust FER system that can deal with particularly challenging images.

The work that is most closely related to our approach is the so-called ExpNet \cite{chang2018expnet}, which uses a CNN to regress 3DMM-based expression coefficients from a single facial image. The proposed approach achieves vastly superior recognition performance (from 21\% to 35\% higher recognition accuracy on 5 different benchmarks, see Table~\ref{tab:results}) with a significantly faster runtime (more than 4 times faster, see Table~\ref{tab:run-time}).
In contrast to ExpNet, our approach adopts a pre-processing step of 2D registration in the image space to a template mean face, see Fig.\ref{fig:framework} middle. This significantly reduces the variability of input images and makes the estimations more robust and reliable \cite{mollahosseini2016going}. 
Furthermore, the 3DMM that we use to model identity variation is the LSFM Model \cite{booth2018large}, which has been trained in 2 orders of magnitude more facial identities than the Basel Face Model \cite{paysan20093d} adopted in ExpNet, achieving much more accurate representation of the 3D shape of human faces \cite{booth2018large}.

\begin{table*}[t!]
    \centering
    \begin{tabular}{p{2cm} |p{3.5cm} |p{2cm} |p{2.5cm} |p{2.5cm} |p{2cm}}
      \textbf{Dataset} &  \textbf{$\#$ images} &  \textbf{$\#$  subjects} &  \textbf{Emotions} &  \textbf{Elicitation} &  \textbf{Resolution} \\
     \hline
       RadFD\cite{langner2010presentation}&  8040&   67&   7 B+ 1 N&  posed &   $681\times1024$\\ 
      \hline
        KDEF\cite{lundqvist1998karolinska}&   4900&   70&   6 B+ 1 N &   posed&   $562\times762$\\
     \hline
       RAF-DB\cite{li2017reliable}&   29672&   N/A&   6 B+ 1 N, 12 C&  posed$\&$ spontaneous&    web images\\
     \hline
       CFEE\cite{du2014compound}&   5060&   230&   6 B+ 1 N, 15 C &   posed&   $1000\times750$\\
     \hline
      CK+\cite{lucey2010extended}&  327 seq. (10 to 60 frames/seq.) &  210 &   7 B+ 1 N &   posed$\&$ spontaneous& $640\times480$  \\
    \end{tabular}
    \caption{Public databases of emotions utilised in this paper. B, N, C stand for basic, neutral and compound emotions, respectively.}
    \label{tab:databases}
\end{table*}

\section{Methodology}
\label{sec:Methodology}
Figure \ref{fig:framework} demonstrates an overview of the proposed framework.  
Motivated by the progress in the 3D facial reconstruction from images and the rich dynamic information accompanying videos of facial performances, we collected a large-scale dataset of facial videos from the internet (section \ref{subsec: dataset}) and recovered the per-frame 3D geometry thereof with the aid of 3D Morphable Models (3DMMs) \cite{blanz1999morphable} of identity and expression (section \ref{subsec: 3DMM}). The annotated dataset was used to train the proposed \textbf{DeepExp3D} network, in a supervised manner for regressing the expression coefficients vector $\mathbf{e}_{f}$ from a single input image $\mathbf{I}_{f}$ (section \ref{subsec: ExpNet}). As a final step, a classifier was added to the output of the DeepExp3D to predict the emotion of each estimated facial expression, and was trained and tested on standard benchmarks for FER ( section \ref{subsec: EmotionClassifier}).


\subsection{3D Face Reconstruction From Videos}
\label{subsec: 3DMM}

\subsubsection{Combined Identity and Expression 3D Face Modelling}
\label{subsec:id_exp}
Following several recent methods \cite{Zafeiriou2017,Deng2018,koujan2018combining,gecer2019ganfit}, we model the 3D face geometry using 3DMMs and an additive combination of identity and expression variation. In more detail, 
let $\mathbf{x}=[x_{1}, y_{1}, z_{1}, ..., x_{N}, y_{N}, z_{N}]^T \in \mathbb{R}^{3N}$ be the vectorized form of a 3D facial shape consisting of $N$ 3D vertices. We consider that any facial shape $\mathbf{x}$ can be represented using the following model of shape variation:
\begin{equation}
\label{eq:3DMM}
\mathbf{x}(\mathbf{i},\mathbf{e})=\bar{\mathbf{x}}+\mathbf{U}_{id} \mathbf{i}+ \mathbf{U}_{exp} \mathbf{e}
\end{equation}
where $\mathbf{\bar{x}}\in \mathbb{R}^{3N}$ is the overall mean shape vector, given by $\mathbf{\bar{x}}=\mathbf{\bar{x}}_{id}+\mathbf{\bar{x}}_{exp}$,  where $\mathbf{\bar{x}}_{id}$ and $\mathbf{\bar{x}_{exp}}$  are the mean identity and mean expression shape vectors respectively. $\mathbf{U}_{id} \in \mathbb{R}^{3N\times n_i}$ is the orthonormal basis with $n_i=157$ principal components ($n_i\ll3N$) , $\mathbf{U}_{exp} \in \mathbb{R}^{3N\times n_e}$ is the orthonormal basis with the $n_e=28$ principal components ($n_e\ll3N$), and $\mathbf{i} \in \mathbb{R}^{n_i}$, $\mathbf{e} \in \mathbb{R}^{n_e}$ are the identity and expression parameters. In the adopted model \eqref{eq:3DMM}, the 3D facial shape $\mathbf{x}$ is a function of both identity and expression coefficients ($\mathbf{x}(\mathbf{i}, \mathbf{e})$). Additionally, the expression variations are effectively represented as offsets from a given identity shape. 

The identity part of the model, $\{\mathbf{\bar{x}}_{id}, \mathbf{U_{id}}\}$, originates from the LSFM \cite{booth2018large} built from approximately 10,000 scans of different people, the largest 3DMM ever constructed, with varied demographic information. 
In addition, the expression part of the model, $\{\mathbf{\bar{x}}_{exp}, \mathbf{U}_{exp}\}$ originates from the work of Zafeiriou et al.~\cite{Zafeiriou2017}, who built it using the blendshapes model of Facewarehouse \cite{cao2014facewarehouse} and adopting Nonrigid ICP \cite{cheng2017statistical} to register the blendshapes model with the LSFM model. 


To create effective pseudo-ground truth, we need to perform 3D face reconstruction on an especially large-scale video dataset that is both efficient and accurate. 
For this reason, we choose to fit the adopted 3DMM model on the sequence of facial landmarks over each video of the dataset.  Since this process is intended for the creation of pseudo-ground truth on a large collection of videos, we are not constrained by the need of online performance. Therefore, we adopt the approach of \cite{Deng2018} (with the exception of the initialization stage, as described next), which is a batch approach that takes into account the information from all video frames simultaneously and exploits the rich dynamic information usually contained in facial videos. This is an energy minimization approach to fit the combined identity and expression 3DMM model on facial landmarks from all frames of the input video simultaneously. We utilise the so-called 3D-aware 2D landmarks which we extract with [12]. The localised 68 landmarks with this method correspond to projections of their corresponding 3D points on the 3D
face. More details are given in the Supplementary Material.

\subsubsection{Initialization Stage of Estimating Camera Parameters} 
In this stage of the 3D video reconstruction proposed in \cite{Deng2018}, the camera parameters are estimated using rigid Structure from Motion (SfM). 
This works reliably for facial videos with substantial head rotation, since it creates the required variation in the relative 3D pose that is typically needed in SfM. However, in cases of videos with almost no or very little head rotation (e.g. a video of a person looking straight at the camera and talking), SfM yields a very unstable estimation of the camera parameters, due to the ambiguities caused when viewing the scene from almost the same view point. To overcome this limitation and exploit much wider types of facial videos, we adopt a substantially different approach in this stage, which utilizes earlier the adopted 3D face model and effectively constraints the problem, yielding not only robust but also accurate estimations. 

In more detail, similar to \cite{Deng2018},  our initialization stage assumes that the shape to be recovered remains rigid over the whole video. This assumption is over-simplistic but is adequate for an accurate estimation of camera parameters, since the deformations in human faces can be reliably modelled as localized deviations from a rigid shape. 
However, in contrast to \cite{Deng2018}, we do not 
seek to estimate the full degrees of freedom of the 3D facial shape (i.e.~every coordinate of every point of the 3D shape being a separate independent parameter);  instead we significantly reduce the allowed degrees of freedom by imposing the constraint that it is synthesised using the 3D face model \eqref{eq:3DMM}. This makes our camera estimations much more robust. Please refer to the supplementary materials for more details about the implementation of this stage. 

\subsection{Ground Truth Creation from a Large-scale  Videos Dataset}
\label{subsec: dataset}

This section describes how we process a very large-scale dataset to construct pseudo-ground truth, which was used to train the \textbf{DeepExp3D}, a robust CNN capable of regressing the 3DMM expression parameters from a single RGB image. We start from a collection of 6,000 RGB videos with 12 million frames in total and 1,700 unique identities. Please refer to the Supplementary Material for the specifics of the collection process.
In every frame of every video of our video collection, we 
applied the method of \cite{deng2018cascade} to detect faces and extract from each detected face a set of 68 landmarks, according to the MULTI-PIE markup scheme \cite{gross2010multi}. Afterwards, we applied the following steps: \\
\textbf{False detections removal:} This was implemented by tracking each detected face in the first frame throughout the processed video. A face is kept if its bounding box (BB) stays within a reasonable margin, chosen experimentally to be half the width of the BB, compared to its location in the previous frame. We pruned videos in which we lost track of the face for $K$ consecutive frames (chosen experimentally to be 5) before reaching the desired number of tracked frames $F$  (chosen experimentally to be 2000). This step helped to remove false detections arising due to a failure in the face detector or out-of-context detections, e.g. a facial photo in the background of a video, faces that pop in/out of the camera viewing angle, etc. 
This step resulted in pruning 1000 videos (16.7\% of the initial dataset).
\\
\textbf{Temporal smoothing:} Extracted landmarks were temporally smoothed using cubic splines. This was performed to alleviate the effects of the potential jitters in the extracted landmarks between consecutive frames and to fill in the possible gaps (frames with lost tracking) that persisted for less than $K$ frames. \\
\textbf{3D facial reconstruction from videos}: For every video, we followed the process described in Sec.~\ref{subsec: 3DMM} and estimated the facial shape parameters ($\mathbf{i}$, $\mathbf{e}_f$ for $f= 1,..., F$). 
The final output of pseudo-ground truth creation is the sequence of expression vectors $\{\mathbf{e}_f\}$. However, we also utilise the identity vector $\mathbf{i}$ in the next step as one means of error pruning.
\\
\textbf{Error pruning}: 
With such a large number of videos, there will be some cases of videos where 3D reconstruction has failed. 
This is an unavoidable byproduct of the fact that the adopted landmark localization, even though very robust, might not be sufficiently accurate for cases of extremely challenging facial videos. 
Our approach compensates for that by two stages of pruning problematic videos: 

\textbf{a) Automatic pruning:} 
We are based on the fact that 
under the adopted 3D face modelling \eqref{eq:3DMM}, 
the coordinates of the estimated identity vector $\mathbf{i}$ of each video 
are assumed independent, identically distributed random variables that follow a normal distribution. 
Therefore, we classify as outliers and automatically prune the videos 
that correspond to an estimated value of $\|\mathbf{i}\|$ above an appropriate threshold.
More details are given in the Supplementary Material. 
This resulted in automatically pruning 300 more videos (5\% of the initial dataset). 








\textbf{b) Manual pruning:}
There might be a few problematic videos that ``survived'' the automatic pruning. For that reason, we inspected the reconstructions of all remaining videos and manually flag and prune videos where it is evident that the 3D face reconstruction has failed. 
In this step we manually pruned 250 videos (around 4\% of the initial dataset).

To conclude, our constructed training set consists of videos of our collection that survived the aforementioned steps of video pruning. It consists of 5,000 videos (83.33\% of the initial dataset) with 1,500 different identities and around 9M frames. For exemplar visualisations, please refer to the Supplementary Material.

\subsection{DeepExp3D Network}
\label{subsec: ExpNet}

The constructed training dataset of videos is rich in the facial expressions that are viewed from different angles and under various illumination conditions throughout the video, as well as in identities (1500 in total). This substantially facilitates the process of training a Convolutional Neural Network (CNN) $\mathcal{N}:I \rightarrow \mathbf{e},$ aiming at regressing the 3DMM facial expression coefficients (referred to as $\mathbf{e}$ in equation \eqref{eq:3DMM}) from a given RGB image $I$. The network $\mathcal{N}(I)$ learns during the training phase how to map from the image space to the facial expression space irrespective of the subject's identity shown in image $I$. This is achievable by the virtue of the utilized facial 3DMM which represents the reconstructed face as a summation of identity and expression parts on top of the model mean face ($\mathbf{\bar{x}}$), see equation \eqref{eq:3DMM}. We extract vectors $\mathbf{e}$ from our dataset as a result of the fitting approach explained in section \ref{subsec: 3DMM} and use them as pseudo annotations for training $\mathcal{N}$ in a supervised manner. However, to avoid teaching $\mathcal{N}$ the exact behaviour of our linear model-based fitting approach for estimating the facial expression parameters, we fine-tune our DCNN ($\mathcal{N}$) on the 4DFAB dataset \cite{cheng20184dfab}. The 4DFAB dataset is a large-scale database of dynamic high-resolution 3D faces (more than 1.8M 3D face) with subjects displaying both spontaneous and posed facial expressions. The ResNet \cite{he2016deep} network structure was selected and trained after replacing the output softmax layer by a linear regression layer of $n_e=28$ neurons. Before starting the training, dataset frames were aligned to a template of size $224 \times 224$ having the 68 points mark-up \cite{sagonas2016300} projected from the mean 3D face $\mathbf{\bar{x}}$ into the image space. In total, the trained DeepExp3D is a mapping: $\mathcal{R}^{224 \times 224} \rightarrow \mathcal{R}^{28}$. Note finally that 70\% of the dataset was used for training and the rest were halved for testing and validation. While training, the network minimises the $\ell_2$ norm error between the output and the ground-truth facial expression parameters.

\subsection{Back-end Emotion Classifier}
\label{subsec: EmotionClassifier}
To classify the generated facial expression vectors $\mathbf{e} \in \mathcal{R}^{28}$ produced by the DeepExp3D network $\mathcal{N}$, Error Correcting Output Codes (ECOC) method \cite{dietterich1994solving} was utilised to solve this 7-class (neutral + six basic emotions) classification problem. ECOC strategy combines multiple binary learners to solve the multi-class classification problem. Our binary learner of choice is Support-Vector Machines (SVM). 
10-fold cross validation with the one-versus-all \cite{nilsson1965foundations} coding scheme were implemented to train/test the emotion classifier. The SVM hyper-parameters were optimized using the Bayesian optimization approach \cite{snoek2012practical}.
68-landmarks were extracted from the images of all the employed emotion datasets in table \ref{tab:databases} and used to register them to the mean face template, as done in section \ref{subsec: ExpNet}.

\section{Experimental Results}
\label{sec:Experimental Results}

In this section, we present extensive qualitative and quantitative evaluations and comparisons of our pipeline, as well as its intermediate steps.


\textbf{Implementation and runtimes}. 
Our method uses the ResNet \cite{he2016deep} CNN structure with 50 layers, implemented in TensorFlow \cite{tensorflow2015-whitepaper}. For both training and testing, we use a machine with Nvidia Tesla V100 GPU and Intel(R) Xeon(R) CPU E5-1660 v4@3.20GHz. 
Our overall FER framework achieves 20ms of total processing time per image (i.e.~50 fps when applied on videos). 
Using the same machine, we also ran methods that solve the same (FER) or closely-related problems (3D shape estimation with disentanglement of identity and expression) using single-image input, see Table \ref{tab:run-time}. We observe that our method is at least 4 to 320 times faster than the other tested methods. 
This is mainly due to the particularly compact and descriptive representation of facial expressions that is achieved in our framework.


\begin{table}[t!]
    \centering
    \caption{Comparison of required run-time to produce estimations of facial expression from a single image. 
    }
    \begin{tabular}{p{1.3cm}|p{0.6cm}|p{1.4cm}|p{1cm}|p{1cm}|p{1cm}}
      \textbf{Method} &  ITW \cite{booth20183d} & SfM3DMM \cite{koujan2018combining} &3DDFA \cite{zhu2016face} &  ExpNet \cite{chang2018expnet} & \textbf{Ours}  \\
     \hline
     \textbf{Time (sec)} & 6.4 & 3.0 & 0.6 & 0.088 & \textbf{0.02}
    \end{tabular}
    \label{tab:run-time}
\end{table}

\subsection{Facial Expression Recognition}

\begin{figure}[t!]
    \centering
    \includegraphics[width=.95\columnwidth]{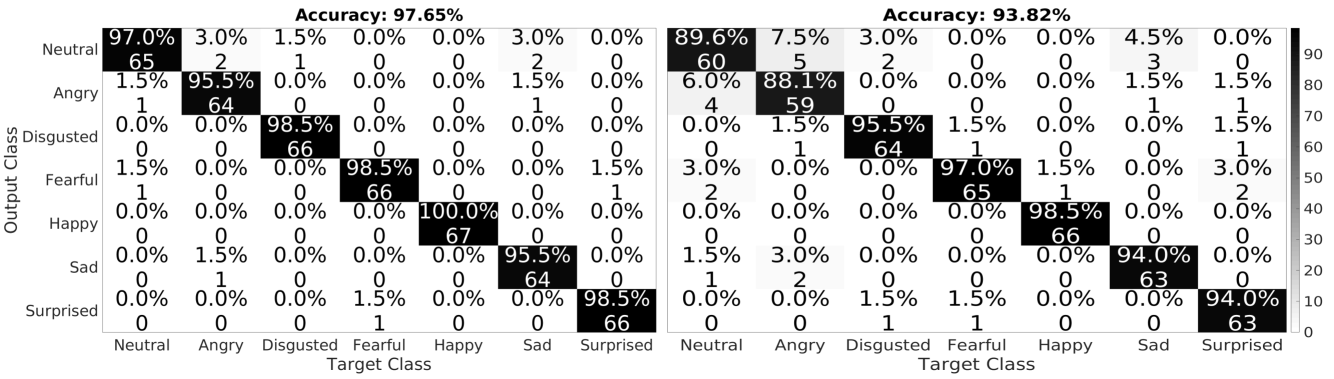}
    \caption{Confusion matrices generated by our emotion classifier when running 10-fold cross validation on RadFD\cite{langner2010presentation} on only frontal images (\textbf{left}) and one of the semi-profile images (\textbf{right}) of each subject. }
    \label{fig:RadFD_ConfMat}
\end{figure}

\begin{figure}[t!]
    \centering
    \includegraphics[width=.95\columnwidth]{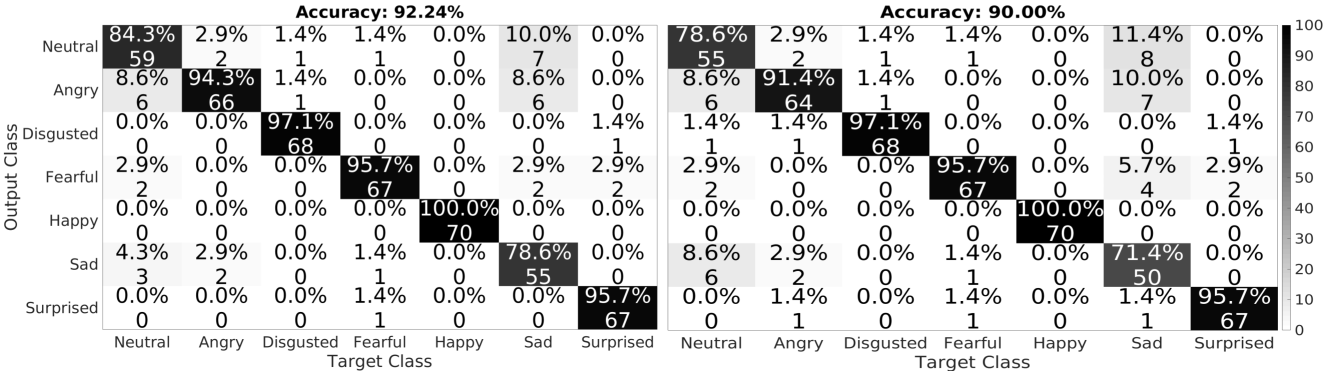}
    \caption{Confusion matrices generated by our emotion classifier when running 10-fold cross validation on the KDEF dataset on only frontal images (\textbf{left}) and only one of the semi-profile images (\textbf{right}) of each subject.}
    \label{fig:KDEF}
\end{figure}

\begin{table*}[!t]
    \centering
    \caption{Comparison of facial expression recognition accuracies on 5 widely-used benchmarks.}
    \begin{tabular}{c c l |c c l |c c l}
     \hline
     \textbf{Dataset} & \textbf{Approach}  &  \textbf{Acc.($\%$)}& 
     \textbf{Dataset} & \textbf{Approach}  &  \textbf{Acc.($\%$)}& 
     \textbf{Dataset} & \textbf{Approach}  &  \textbf{Acc.($\%$)}\\
     \hline
      \multirow{9}{*}{\textbf{RadFD}} &  ExpNet \cite{chang2018expnet}&  75.00 & \multirow{10}{*}{\textbf{RAF-DB}} & ExpNet \cite{chang2018expnet}&  55.20  
      &
       \multirow{6}{*}{\textbf{KDEF}} &   ExpNet \cite{chang2018expnet}&   71.00
      \\
                                                           &Ali et al. \cite{ali2017approach}  & 85.00 &  & Li$\&$ Deng \cite{li2019reliable}&  74.20   & &Zavarez et al. \cite{zavarez2017cross} &    72.55\\
                                                           &   Zavarez et al. \cite{zavarez2017cross} &   85.97 &   & Lin et al. \cite{lin2018facial}&           75.73 & & Ali et al. \cite{ali2017approach} & 78.00\\
                                                           &    Jiang$\&$ Jia \cite{jiang2016robust} &   94.52  &   & Fan et al. \cite{fan2018multi}&  76.73 & & Ruiz-Garcia et al. \cite{ruiz2017stacked} &   86.73 \\
                                                           &    Mavani et al. \cite{mavani2017facial} &   95.71  &  & Gosh et al. \cite{ghosh2018automatic}&      77.48 & & Yaddaden et al. \cite{yaddaden2018user} &  90.62\\
                                                           &    Wu$\&$ Lin \cite{wu2018adaptive} &   95.78  &   & Shen et al. \cite{shen2018double}&          78.60 & & \textbf{Ours}& \textbf{92.24}$\pm$\textbf{0.70}\\
                                                           \cline{7-9}
                                                           &    Sun et al. \cite{sun2017efficient} &   96.93 &  & Vielzeuf et al. \cite{vielzeuf2018occam} & 80.00 & \multirow{4}{*}{\textbf{CK+}} & ExpNet \cite{chang2018expnet} & 61.17\\
                                                           &Yaddaden et al. \cite{yaddaden2018user}  &  97.57$\pm$1.33 &  & Deng et al. \cite{8606936}&     81.83 & &  Wang et al. \cite{wang2013capturing} &  86.3\\
                                                           &    \textbf{Ours}&\textbf{97.65}$\pm$\textbf{1.00}&   &  Ours & 82.06$\pm$0.73 & & Jung et al.    \cite{jung2015joint} & 92.35\\
                                                           \cline{1-3}
                                                         \multirow{3}{*}{\textbf{CFEE}}  &  ExpNet \cite{chang2018expnet}   &   62.50 &  & \textbf{Li et al.} \cite{li2018patch}& \textbf{83.27} & & \textbf{Ours}& \textbf{96.45}$\pm$ \textbf{0.8}\\
                                                          & Ours & 96.43$\pm$ 1.1 & & & & & &\\
                                                          & \textbf{Du et al.}\cite{du2014compound}& \textbf{96.84}$\pm$\textbf{9.73}  & & & & & &\\
    \hline

    \end{tabular}
    \label{tab:results}
\end{table*}

To evaluate our FER method, 5 publicly available datasets for emotion recognition were used, namely: Radboud \cite{langner2010presentation}, KDEF \cite{lundqvist1998karolinska}, RAF-DB \cite{li2017reliable}, CFEE \cite{du2014compound}, CK+ \cite{lucey2010extended}. All five datasets have basic emotion \cite{ekman1993facial} annotations (happy, sad, fearful, angry, surprised, disgusted), as well as the neutral expression. Presentation of results per dataset follows:

First of all, the \textbf{Radboud dataset} \cite{langner2010presentation} has 67 subjects imaged from 5 different angles each at the same time. To test the performance of our network on recognising an emotion from dissimilar view angles, we run two experiments. In the first, the frontal image of each subject showing a specific emotion was kept ($67\times7=469$ images in total), while in the second experiment one of the semi-profile faces (captured from 45/135 degrees) of each subject was selected randomly and used for 10-fold cross validation. Figure \ref{fig:RadFD_ConfMat} reports the confusion matrices and accuracies obtained in both cases. The average MSE of expression parameters generated from semi-profile and frontal images is 0.008 over all the subjects. The comparable generated accuracies in figure \ref{fig:RadFD_ConfMat}, as well as the small MSE, demonstrate the ability of the DeepExp3D in producing view-angle independent expression estimations. As shown in table \ref{tab:results}, our proposed approach produces the highest \textbf{accuracy} (97.63$\%$) on the RadFD \cite{langner2010presentation} dataset compared to recent state-of-the-art methods.

The \textbf{KDEF  dataset} \cite{lundqvist1998karolinska} is similar in structure to RadDF \cite{langner2010presentation} where each of the 70 subjects was pictured from five different angles at the same time ($0^{\circ}$, $45^{\circ}$, $90^{\circ}$, $135^{\circ}$, $180^{\circ}$). Each subject was asked to elicit the same emotion twice, only one thereof was picked randomly. Only frontal images ($7\times70=490$ in total) were employed in the reported results in table \ref{tab:results}. We attain the best \textbf{accuracy} compared to other state-of-the-art methods ($92.24\%$), revealing the power of our DeepExp3D in generating separable facial expressions according to their basic emotion label. Figure \ref{fig:KDEF} demonstrates the confusion matrices generated by our emotion classifier on either frontal images (left) or semi-profile ($45^\circ$) images (right). Both reported confusion matrices emphasize the high separability of happy and disgusted labels from the rest of the emotions ($100\%$ and $97.1\%$, respectively), while it seems that the sad and neutral expressions tend to group closely ($78.6\%$ and $84.3\%$ for frontal, and  $71.4\%$ and $78.6\%$ for semi-profile images, respectively).

The \textbf{CFEE dataset}  \cite{du2014compound} was collected from 230 subjects with two groups of labelled images, basic and compound. Images labelled with basic emotions (total of 1836) were passed to the DeepExp3D and then for training/testing our emotion classifier. Our obtained \textbf{average accuracy per class} is comparable to the state-of-the-art on this dataset by Du et al. \cite{du2014compound}, see table \ref{tab:results}. 

The \textbf{RAF benchmark}  \cite{li2017reliable}  is the most challenging among all utilised FER datasets in this paper. This dataset was collected from the internet, no lab-controlled conditions. The authors of \cite{li2017reliable} sought the help of well-trained annotators for segregating the dataset into basic and compound emotion images. We use the basic emotion images, which are 13395 in total, and estimate their facial expressions using DeepExp3D. The train/test splits provided by the authors of \cite{li2017reliable} were used for training/testing our emotion classifier. Our produced \textbf{average accuracy per class}  is comparable to the highest  accuracy reported on this dataset ($82.06$ vs $83.27$).

On the \textbf{Extended Cohn-Kanade (CK+)} \cite{lucey2010extended} dataset, our method manages to generate the highest accuracy (96.45$\%$) compared to other methods. This dataset has 327 sequences of frontal images originating from 210 subjects. Similar to \cite{chang2018expnet}, we keep the peak frame of each sequence and associate it with the label of this sequence.

\textbf{Overall}, quite similarly in all experimented benchmarks, the trained emotion classifier recognises the happy, fearful, surprised and disgusted emotions better than the rest (neutral, sad, angry). This can be mainly referred to two essential factors: 1) intensity of the related action units when deconstructing each emotion according to the Emotional Facial Action Coding System (EFACS) \cite{friesen1983emfacs}, 2) ability of employed expression basis ($U_{exp}$) in capturing the relevant action units. The trained DeepExp3D tends to capture well mouth- jaw- and cheeks-related motions (action units 6, 12, 14, 15, 16, 20, 26 \cite{friesen1983emfacs}), e.g. lip corner puller/depressor, lower lip depressor, lip stretcher, jaw drop, etc., which are essential in characterising the happy, surprised, disgusted and fearful emotions. On the other hand, subtle details around the eyes, like inner brow raiser, brow lowerer, upper lid raiser, lid tightener, etc., which are judgmental for discerning emotions like sadness and anger, appear to be more challenging for the DeepExp3D. This can be explained by the fact that action units 6, 12, 14, 15, 16, 20, 26 are better represented in the original FaceWarehouse model utilised to annotate our collected dataset of videos presented in section \ref{subsec: dataset}, as well as the 4DFAB dataset \cite{cheng20184dfab} used for the fine-tuning stage. Please see the supplementary materials for more results and visualisations. Additionally, the extracted landmarks might degrade the results largely if they fail to annotate the 68 targeted locations on the face with good accuracy. 

\subsection{Emotional Stress Analysis}
In this section, we investigate the ability of our proposed framework in detecting stress conditions using only facial videos. Stress is widely conceived as a complex emotional state which can be identified by biosignals \cite{ggianreview2019}. However, their recording may not always be convenient and practical for daily monitoring, thus research community pursuits stress identification only using facial cues, which constitutes a quite challenging task. Related literature is limited regarding the combination of biosignals with deep learning frameworks \cite{hwang2018deep, giannakakisnovel}, or only visual cues \cite{giannakakis2018evaluation, giannakakis2017stress}. In this work, we evaluate the performance of our method against other state-of-the-art methods in stress identification. Towards that end, we utilize the dataset (SRD'15) used in \cite{giannakakis2018evaluation} which has 24 subjects (with age 47.3$\pm$9.3 years) and 288 videos in total. Each subject performed 11 experimental tasks (either neutral or stressful). The whole experiment was divided into 4 phases: 1) social exposure, 2) Emotional recall, 3) stressful images/mental task, 4) stressful videos.  

The frames of each recorded video were labeled as either 'stressful' or 'non-stressful', according to the task under investigation. Next, our method was used to perform facial expression recognition from each frame and a 5-fold cross validation was carried out, while making sure frames coming from the same subject do not exist in both training and testing folds at the same time. The experiments were repeated 10 times and the average stress recognition accuracy of each phase is reported in table \ref{tab:stress}. Note that the first phase (social exposure) was not taken into account as it contains a task with speech which affects `per se' head motility compared to a neutral non speech task as explained in \cite{ggianbhi2018}. 
For comparisons, we have also followed the same protocol and applied the method of \cite{giannakakis2018evaluation} which uses head motility and 
the method of \cite{giannakakisnovel} which uses heart activity signals (IBI), with their results also shown in table \ref{tab:stress}.  
We observe that the proposed method achieves high accuracy and outperforms the other tested methods. 
This is an especially promising result for stress analysis, as our method uses only non-invasive and frame-based visual features.

\begin{table}[!t]
    \centering
    \caption{Stress detection accuracy comparison on dataset used in \cite{giannakakis2018evaluation}}
    \begin{tabular}{c |c |c |c}
     \textbf{Phase}  &  \textbf{Head motility} \cite{giannakakis2018evaluation}& \textbf{DWNet1D} \cite{giannakakisnovel} &  \textbf{Ours} \\
     \hline
     Emotional Recall  &82.99 \% & 83.50 \% & \textbf{86.70 \%}\\
     \hline
     Stressful images  &85.42 \%& \textbf{92.60 \%} & 88.42 \%\\
     \hline
     Stressful videos &85.83 \%& 85.90 \% & \textbf{88.83 \%}\\
     \hline
     \textbf{Average}&  84.75 \%& 87.33 \% &  \textbf{87.98 \%}\\
    \end{tabular}
    \label{tab:stress}
\end{table}

\subsection{Evaluation of our Framework's Intermediate Steps}
\begin{figure}[t!]
    \centering
    \includegraphics[width=\columnwidth, trim=0 0 480 0, clip]{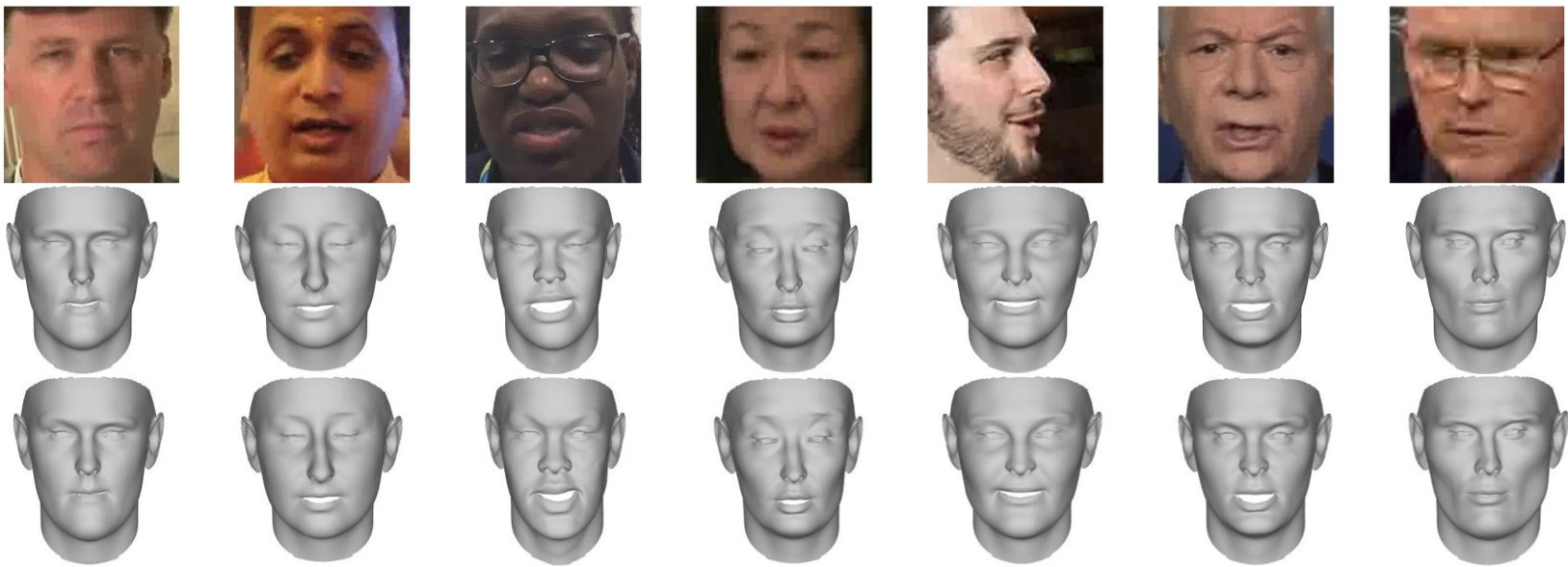}
    \caption{Estimated expressions (second row) from selected test images (first row) of our dataset with the Ground Truth (GT) expressions (bottom row). Both estimated and GT expressions are visualized with the 3D GT identity. }
    \label{fig:tmf2_test}
\end{figure}

Even though the final output of our proposed pipeline is the emotion class, we 
have also conducted detailed experiments to evaluate the intermediate steps of our framework.
First of all, we qualitatively and quantitatively evaluate the accuracy of estimating the \textbf{3D expression parameters}, which is the intermediate pipeline step taken as output of DeepExp3D. 
We test its performance on the test split of our \textbf{FaceVid} dataset both qualitatively and quantitatively. First of all, Figure \ref{fig:tmf2_test} presents qualitative results of the proposed method as compared to the Ground Truth (GT) reconstructions. We show both the estimated and GT expressions on top of the mean face and the GT identity parameters. 
We observe that the estimations provided by our method are visually very close to the GT. 
Furthermore, we compare our DeepExp3D with: 1) a baseline approach following a linear shape model fitting proposed in  \cite{huber2017real}, and 2) ITW, a state-of-the-art 3D reconstruction method from in-the-wild images \cite{booth20183d}. We provide both methods with the same facial expression model (FaceWarehouse \cite{cao2014facewarehouse}) and compute the average of the Mean Squared Error (MSE) between the estimations and the ground truth over the test split. Our method achieves by far the lowest (best) MSE with \textbf{0.007}, while ITW \cite{booth20183d} and the baseline \cite{huber2017real} obtain 0.026 and 0.098, respectively. More results and visualisations are available in the supplementary material.

Furthermore, we evaluate our approach on \textbf{3D face reconstruction from videos} (Sec.~\ref{subsec: 3DMM}), which is used to construct the pseudo-ground truth annotations of our training dataset (\textbf{FaceVid}). 
This evaluation is presented in the Supplementary Material, from which it can be concluded that our proposed approach outperforms the compared state-of-the-art 3D reconstruction methods and achieves satisfactory accuracy for usage in pseudo-ground truth creation.

\section{Conclusion}
\label{sec:Conclusion}
In this paper, we have proposed a framework for the automatic recognition of human emotions from monocular images. Our framework utilises a well-trained deep CNN (termed as \textbf{DeepExp3D}), of our own implementation, capable of estimating the 3D facial expression parameters from a single image, even in challenging scenarios.  We have extensively evaluated the performance of our trained \textbf{DeepExp3D} and compared it with state-of-the-art methods for 3D reconstruction from in-the-wild images. Our \textbf{DeepExp3D} demonstrates a superior performance in regressing the facial expression coefficients when trained on the same facial expression model as the competitors. We have also extensively tested the potential of the trained DeepExp3D in recognising facial expressions when combined with an mSVM classifier on 5 widely-used benchmarks (taken under either controlled or in-the-wild conditions). Our reported emotion recognition results reveal the competitive performance of our proposed framework when compared with recent state-of-the-art approaches on the same datasets. We report the highest accuracy on the KDEF \cite{lundqvist1998karolinska} ($92.24\%$), RadFD \cite{langner2010presentation} ($97.63\%$), CK+ \cite{lucey2010extended} ($96.45\%$) datasets, and the second best on CFEE \cite{du2014compound} and RAF-DB \cite{li2017reliable} (with $0.41\%$ and  $1.21\%$ difference from the best, respectively).

{\small
\bibliographystyle{ieee}
\bibliography{egbib}
}







\end{document}